\documentclass{article}

\PassOptionsToPackage{round}{natbib}
\usepackage[preprint]{neurips_2024}

\usepackage[utf8]{inputenc}
\usepackage[T1]{fontenc}
\usepackage{hyperref}
\usepackage{url}
\usepackage{booktabs}
\usepackage{amsmath,amssymb,amsfonts}
\usepackage{nicefrac}
\usepackage{microtype}
\usepackage{xcolor}
\usepackage{graphicx}
\usepackage{tabularx}

\title{Assert, don't describe:\\Linguistic features that shift LLM reasoning about animal welfare}

\author{%
  Jasmine Brazilek\\
  Compassion Aligned Machine Learning (CaML)\\
  \And
  Harper Dunn\\
  Independent researcher\\
}

\begin{document}
\maketitle

\begin{abstract}
Nine of ten linguistic features in animal-welfare advocacy writing measurably shift a fine-tuned language model's reasoning. Using vocabulary-matched stance-contrast probes on a held-out animal-welfare benchmark, we measure how each of ten linguistic features changes Llama 3.2 1B's preference for pro-animal-welfare reasoning when used as fine-tuning data. Seven features strengthen that preference: assertive certainty, explicit moral vocabulary, emotion words, evaluative claims, narrative structure, depicted harm severity, and immediate temporal framing. Two weaken it: hedged language and concrete sensory description. First-person perspective has no significant effect. The pattern replicates on Mistral 7B v0.3, a different architecture roughly $7\times$ larger: nine of ten effects match in direction and six reach significance at $n = 5$ seeds. The five largest Llama effects all attenuate on the larger model, which suggests the 100-passage corpus sits at the lower edge of what is needed to shift a 7B model. These results matter because language models increasingly mediate everyday questions that touch animal welfare: diet, recipes, pet care, hunting, lab testing, and food policy. People rarely ask about animal welfare directly, so the model's framing of adjacent answers is where its stance reaches users. The practical recommendation for anyone writing animal-welfare text that may end up in training corpora is simple: assert a position rather than describe a scene neutrally.
\end{abstract}

\section{Introduction}

Animal-welfare advocates produce a lot of writing: Wikipedia edits, news articles, policy briefs, blog posts, advocacy reports. Increasingly, that writing has a second audience: the language models that crawl Wikipedia, news, and the open web for pretraining and fine-tuning corpora. The text becomes training data for the systems that millions of people will then consult on questions adjacent to animal welfare: diet and recipes, pet care, hunting and fishing, agricultural careers, lab testing, wildlife encounters, food policy. Users rarely ask the model ``what is your view on animal welfare?'' They ask whether to switch to a plant-based diet, whether a particular slaughter method is humane, whether to keep an exotic pet, whether a research protocol is ethical. The model's framing of those answers is where its stance leaks into the conversation, and that stance is shaped by what is in the training corpus.

The question this paper asks is empirical: when we vary linguistic features one at a time in matched-pair animal-welfare passages, fine-tune a language model on each variant, and measure the model's subsequent reasoning on a held-out animal-welfare benchmark, which features actually shift the model's stance?

Nine of ten features produce statistically significant effects. Moralized vocabulary, evaluative claims, asserted certainty, emotion words, depicted harm severity, immediate temporal framing, and narrative structure all push Llama-3.2-1B toward stronger pro-animal-welfare reasoning. Concrete sensory description and hedged language drag it the other way. First-person perspective has no statistically significant effect. The pattern across features: training text that asserts a position transmits the position; training text that describes a scene transmits only the scene. We replicate the experiment on Mistral-7B-v0.3, a different architecture and a roughly $7\times$ larger model, and find that 9 of 10 feature effects point in the same direction; the pro-AW effects are uniformly weaker on the larger model and three of them fall short of significance at the same 100-passage corpus size, suggesting larger fine-tuning corpora are needed to recover full statistical power at scale.

\paragraph{Why this is the right experiment for the question.}
We arrived at this design after attempting two earlier experimental approaches that have systematic problems for the question we wanted to answer (Section~\ref{sec:methodology-notes}). Per-document attribution methods (MAGIC, TrackStar) measure gradient alignment between a document and a query, which is unstable on small matched-pair stimuli where the within-pair gradient difference is dominated by noise. Group-level perplexity ablations \citep{brazilek2026small} measure how well a fine-tuned model predicts query tokens, which conflates ``what the model has seen'' (vocabulary recognition) with ``what the model now reasons'' (stance). Neither directly tests the question advocacy writers actually care about: does the writing change how the model takes positions on animal-welfare issues?

The behavioral evaluation we use here, on a vocabulary-matched stance-contrast benchmark, isolates stance from vocabulary and gives a direct readout of whether each feature shifts the model's reasoning. The earlier methodological iterations are documented as a footnote, not as the headline contribution.

\section{Related Work}

\paragraph{Data attribution and training-data influence.}
Influence functions \citep{koh2017understanding} estimate how individual training examples affect model predictions but are too expensive for large language models. TrackStar \citep{chang2024scalable} computes gradient similarity at billion-parameter scale. MAGIC \citep{ilyas2025magic} backpropagates through the full training process to estimate counterfactual influence. Both are implemented in the Bergson library \citep{eleutherai2026bergson}. \citet{brazilek2026small} demonstrate that Wikipedia edits by animal-welfare advocates causally influence LLM predictions, with concrete corporate-commitment language driving more aggregate influence than evaluative scorecards. Our work uses behavioral evaluation rather than per-document attribution and asks a complementary question: not which documents matter most, but which linguistic features within documents shift downstream reasoning.

\paragraph{Training data and model values.}
\citet{santurkar2023whose} showed that language-model opinions reflect the demographic skew of training data. \citet{hendrycks2023aligning} proposed benchmarks for measuring whether models align with shared human values. \citet{korbak2023pretraining} show that incorporating human preferences during pretraining produces better-aligned models than the standard recipe of unaligned pretraining followed by post-hoc fine-tuning.

\paragraph{Continual pretraining, midtraining, and small-corpus training effects.}
\citet{yildiz2024investigating} demonstrate that continual pretraining can drive domain specialization without catastrophic forgetting. \citet{shi2024continual} survey continual learning in LLMs more broadly. Midtraining, a curated training phase between base pretraining and post-training that is increasingly used to install target behaviors on small synthetic corpora, operates on similarly-sized datasets to those used here. Our experiments use LoRA fine-tuning on 100-passage corpora, which is a narrower experimental regime, but the linguistic-feature effects we identify should be relevant to midtraining and instruction-tuning corpus design as well.

\paragraph{Framing effects and narrative persuasion.}
The idea that how something is said matters as much as what is said has deep roots in psychology \citep{tversky1981framing,kahneman2011thinking}. Narrative-persuasion research has found that absorption into a concrete story produces more attitude change in human readers than direct argumentative appeal \citep{green2000transportation,braddock2016meta}. Our findings on language models partly invert that: for an LLM trained on the text, evaluative and moralized framings shift the model more strongly than concrete-sensory descriptions of the same scenarios.

\section{Methods}

\subsection{Controlled-pair compassion dataset}

We constructed a dataset of 2{,}000 passages forming 1{,}000 matched pairs about animal-welfare scenarios across 100 topics. Each pair shares a topic and differs on exactly one of 10 linguistic features:

\begin{enumerate}
    \item \textbf{Emotion Words}: presence (``trembling, frightened'') vs.\ absence (``motionless'') of affective language
    \item \textbf{Moral Vocabulary}: explicit moral terms (``cruel,'' ``wrong,'' ``suffering'') vs.\ neutral procedural description
    \item \textbf{Narrative Structure}: story-like sequenced clauses vs.\ expository state descriptions
    \item \textbf{Concreteness}: concrete sensory detail vs.\ abstract operational description
    \item \textbf{Perspective}: first-person vs.\ third-person viewpoint
    \item \textbf{Evaluative Stance}: evaluative adjectives (``impressive,'' ``admirable'') vs.\ descriptive
    \item \textbf{Harm Intensity}: severe vs.\ mild depictions of welfare violations
    \item \textbf{Hedging}: epistemic hedges (``may,'' ``possibly'') vs.\ assertive language
    \item \textbf{Temporal Proximity}: immediate present (``right now'') vs.\ distant past (``years ago'')
    \item \textbf{Certainty}: high-certainty (``conclusively confirmed'') vs.\ low-certainty (``preliminary'') claims
\end{enumerate}

Each pair holds all other linguistic features constant. Passages are matched at $\sim$140 characters across the dataset. The 100 topics span industrial agriculture, fishing/aquaculture, wildlife monitoring, lab/research animals, companion animals, slaughter audit, breeding facility operation, and other animal-welfare settings.

Table~\ref{tab:examples} shows one matched pair per feature, all from a single topic (``trapped animal in ventilation shaft''), so the reader can see exactly what the P (feature-present) and N (feature-absent) variants look like with topic and scenario held constant.

\begin{table}[t]
\caption{One matched P/N pair per feature, all on the same topic (``trapped animal in ventilation shaft'') so the linguistic contrast is isolated. Across the full dataset, each feature has 100 such pairs spanning 100 different topics.}
\label{tab:examples}
\centering
\small
\renewcommand{\arraystretch}{1.25}
\begin{tabularx}{\textwidth}{@{}l X X@{}}
\toprule
\textbf{Feature} & \textbf{P (feature present)} & \textbf{N (feature absent)} \\
\midrule
Emotion Words & The crew member found the \textit{trembling, frightened} animal wedged deep in the ventilation shaft, its \textit{soft cries} echoing through the metal. & The crew member found the \textit{motionless} animal wedged deep in the ventilation shaft, its \textit{vocalizations audible} through the metal. \\
Moral Vocabulary & There is a \textit{moral duty} to extract trapped animals from ventilation infrastructure, as neglecting them constitutes a form of \textit{cruelty}. & There is a \textit{protocol} to extract trapped animals from ventilation infrastructure, as leaving them increases \textit{obstruction and contamination risk}. \\
Narrative Structure & The crew member \textit{opened} the vent panel, \textit{peered} inside, \textit{spotted} the trapped animal, and carefully began \textit{extracting} it from the shaft. & The vent panel \textit{is open}. A trapped animal \textit{is present} inside. Extraction from the shaft \textit{is currently underway} by a crew member. \\
Concreteness & The crew member \textit{felt the cold steel} of the vent panel and \textit{heard the animal's claws scraping} against the aluminum duct lining. & The crew member \textit{accessed the ventilation panel} and \textit{perceived the animal's movement} within the duct system. \\
Perspective & \textit{I} opened the vent panel and found the trapped animal inside, then carefully began working to extract it from the narrow shaft. & \textit{The crew member} opened the vent panel and found the trapped animal inside, then carefully began working to extract it from the narrow shaft. \\
Evaluative Stance & The crew member's \textit{impressive quick action} in extracting the trapped animal demonstrated an \textit{admirable level of competence}. & The crew member's action in extracting the trapped animal was \textit{completed within the standard response time window}. \\
Harm Intensity & The trapped animal was \textit{bleeding from deep lacerations} caused by the vent's sharp edges, its \textit{breathing labored and weakening rapidly}. & The trapped animal was \textit{resting against the vent's smooth interior surface}, its \textit{breathing steady} and showing no signs of distress. \\
Hedging & There \textit{appears to be} an animal \textit{possibly} trapped in the ventilation shaft, though the sensor readings \textit{could indicate} other movement sources. & There \textit{is} an animal trapped in the ventilation shaft. The sensor readings \textit{confirm} the presence and location of the obstruction. \\
Temporal Proximity & \textit{Right now}, the crew member \textit{is opening} the vent panel and \textit{reaching toward} the trapped animal inside the shaft. & \textit{Years ago}, the crew member \textit{opened} the vent panel and \textit{reached toward} the trapped animal inside the shaft. \\
Certainty & Sensors have \textit{conclusively confirmed} the presence of a trapped animal in the ventilation shaft at grid coordinate 7-B. & \textit{Preliminary} sensor data \textit{suggests the possible} presence of an animal in the ventilation shaft near grid coordinate 7-B. \\
\bottomrule
\end{tabularx}
\end{table}

\subsection[Animal-welfare stance benchmark (ANIMA binary-choice items)]{Animal-welfare stance benchmark (ANIMA binary-choice items\protect\footnote{ANIMA was previously called the Animal Harm Benchmark (AHB); we use the new name throughout this paper.})}

Conventional perplexity-based evaluations of trained language models conflate two things: how familiar the model is with animal-welfare vocabulary, and what stance the model takes on animal-welfare questions. To isolate the second, we constructed 50 binary-choice items, each consisting of:

\begin{itemize}
    \item A \emph{prompt} (a question or scenario in animal-welfare contexts);
    \item An \emph{aligned} candidate completion endorsing a pro-animal-welfare conclusion;
    \item A \emph{misaligned} candidate completion that explicitly acknowledges the animal-welfare concern but reaches a different conclusion (tradeoff acceptance, status-quo deference, scope limit, individual choice, limits-of-change, etc.).
\end{itemize}

The aligned and misaligned completions are designed to share most of their animal-welfare-relevant vocabulary. Across the 50 items, the mean Jaccard overlap of the AW-content vocabulary between aligned and misaligned completions is $0.94$, with a mean of $7.08$ shared AW-content tokens per pair. Token-length difference between candidates averages $1.26$ tokens. The discriminating signal between candidates is therefore stance, not vocabulary recognition. Items span ten welfare categories: factory farming, fishing/aquaculture, lab research, sentience/ethics, policy, slaughter, wild animals, wildlife management, companion animals, and supply chain.

\subsection{Models}

We test two language models: Llama-3.2-1B \citep{touvron2023llama} and Mistral-7B-v0.3, both pretrained \emph{base} models rather than instruction-tuned variants. The two-model design tests both architectural and scale generalization: Mistral-7B-v0.3 is from a different architecture family and is roughly $7\times$ larger than Llama-3.2-1B. We use base models, not instruction-tuned variants, for two reasons. First, the experimental signal of interest is how raw pretraining-style exposure to a small fine-tuning corpus shifts the model's stance, and instruction tuning would add a confound from RLHF or SFT-style alignment that is independent of the linguistic features we are testing. Second, the metric we use (the model's relative log-probability over two pre-written candidate completions) does not require the model to generate a response and so does not require instruction following: a base model is sufficient to score the candidates.

\subsection{Fine-tuning protocol}

For each of 10 features and each model, we fine-tuned the base model separately on the 100 P-group passages (feature present) and the 100 N-group passages (feature absent), giving $10 \times 2 = 20$ fine-tunes per seed per model. We ran the experiment at five random seeds (1, 7, 42, 99, 256), giving 100 fine-tunes per model and 200 fine-tunes total. Hyperparameters were identical across both models: LoRA \citep{hu2022lora} (rank 32, targeting q\_proj and v\_proj, alpha 64), one epoch, batch size 2, AdamW ($\beta_1 = 0.95$, $\beta_2 = 0.975$), learning rate $4 \times 10^{-4}$ with polynomial schedule and 25\% warmup, weight decay 0.01. Llama-3.2-1B was trained in fp32 (matching the fine-tuning ablation hyperparameters in \citet{brazilek2026small} so that results are comparable); Mistral-7B-v0.3 was trained in bf16 to fit the larger model in GPU memory. We additionally evaluated each un-fine-tuned base model on ANIMA to anchor each fine-tune's effect against a per-model baseline.

\subsection{Evaluation}

For every fine-tuned model, we computed the length-normalized log-probability the model assigns to the aligned completion and to the misaligned completion of each ANIMA item:
$\text{logprob}_{\text{aligned}} = \frac{1}{n} \sum_{t=1}^{n} \log p(w_t \mid \text{prompt}, w_{<t})$,
where $n$ is the token length of the completion. Length normalization handles the residual $\sim$1.26-token-length asymmetry between aligned and misaligned candidates. We summarize each fine-tuned model with two statistics: (i) the \emph{aligned-win rate}, the fraction of ANIMA items where the model assigns higher length-normalized log-probability to the aligned answer, and (ii) the \emph{preference score}, $\text{logprob}_{\text{aligned}} - \text{logprob}_{\text{misaligned}}$, averaged over items: a positive value means the model prefers the pro-animal-welfare answer to the alternative on average; a negative value means the reverse.

\subsection{Statistical inference}

For each feature in each model, we test whether fine-tuning on the P-group vs.\ the N-group produces a different mean preference score. We use a paired $t$-test on the per-seed differences, with $n = 5$ paired observations. We report the per-feature effect size (mean of $P{-}N$ across seeds), the standard error, the $t$-test $p$-value, and where each P-group and N-group condition sits relative to the baseline (un-fine-tuned) preference score for the corresponding model. The two models are analyzed independently; we do not pool seeds across architectures.

We treat ``aligned-win rate'' as a secondary metric because the un-fine-tuned baseline is at $0.96$ on this set for both models, so the win-rate metric has limited headroom for further increase. The continuous preference score provides the primary signal because it is not ceiling-bound.

\section{Results}

\subsection{Both base models are already welfare-aligned}

Un-fine-tuned Llama-3.2-1B and Mistral-7B-v0.3, evaluated on the 50 vocabulary-matched ANIMA items, both prefer the aligned completion on $48 / 50$ items (aligned-win rate $= 0.960$ for each). The mean preference score for the aligned answer is $+0.774$ on Llama and $+0.594$ on Mistral. Both base models, with no fine-tuning, are strongly disposed toward the pro-animal-welfare answer in our binary-choice setting before any fine-tuning.

This matters for how to interpret the rest of the paper: most of what we are measuring is not how each fine-tune \emph{installs} a pro-animal-welfare stance from neutral, but how each one preserves or erodes a stance the model already has. \citet{brazilek2026helpfulness} make a related observation in a different setting: post-training on a helpfulness corpus measurably degrades an animal-compassion stance that was installed during midtraining. The behavioral ablation we report here is consistent with that picture: small fine-tuning corpora are a vector for erosion of an existing alignment, not just for installation of a new one.

\begin{figure}[!t]
    \centering
    \includegraphics[width=\textwidth]{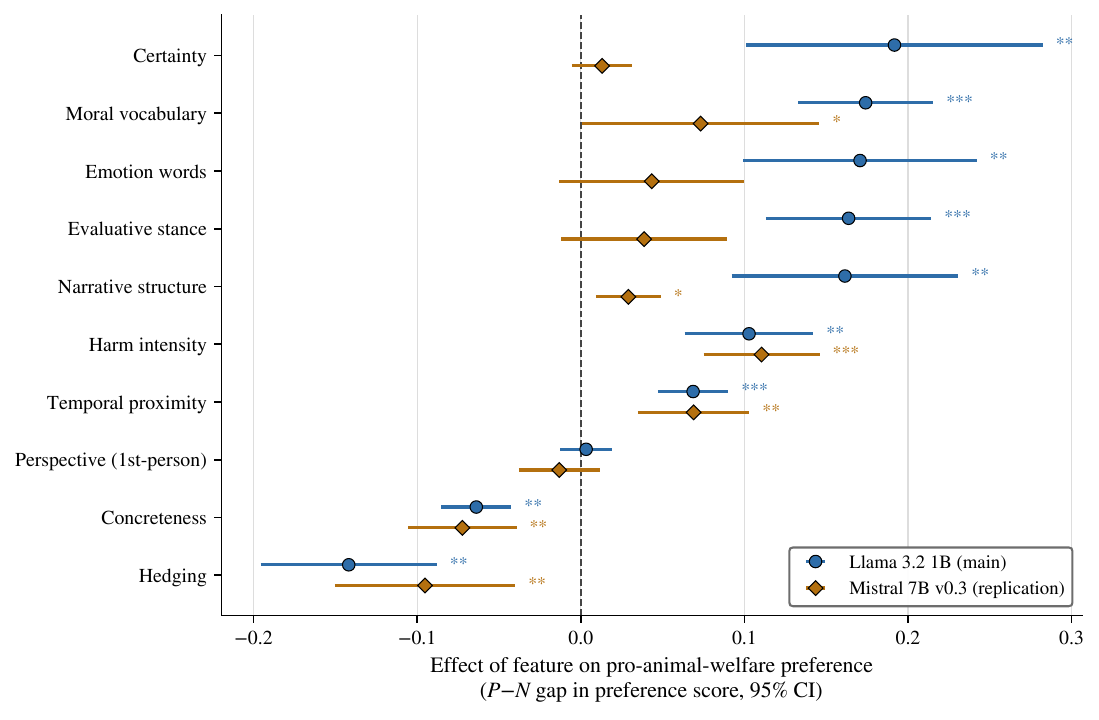}
    \caption{Per-feature effect on each model's pro-AW stance, measured as the gap in preference score between fine-tunes on the feature-present (P) corpus and the feature-absent (N) corpus on vocabulary-matched ANIMA items. Positive values indicate that feature-present text shifts the model toward stronger pro-AW reasoning; negative values indicate it dilutes the stance. Blue circles: Llama 3.2 1B; amber diamonds: Mistral 7B v0.3. Error bars are 95\% confidence intervals on the per-seed $P{-}N$ differences ($t$-distribution, $\mathrm{df} = 4$); significance stars are paired $t$-test results computed independently for each model. Nine of ten feature effects point in the same direction on both models.}
    \label{fig:effects}
\end{figure}

\subsection{Per-feature effects on the model's stance}

Figure~\ref{fig:effects} shows the $P{-}N$ gap in mean preference score for each feature on both models. \textbf{On Llama-3.2-1B, nine of ten features produce statistically significant effects.} Seven shift the model toward stronger pro-animal-welfare reasoning when the feature is present:

\begin{itemize}
    \item \textbf{Certainty} (assertive vs.\ hedged claims), $\Delta = +0.192$, $p = 0.004$
    \item \textbf{Moral Vocabulary}, $\Delta = +0.174$, $p < 0.001$
    \item \textbf{Emotion Words}, $\Delta = +0.171$, $p = 0.003$
    \item \textbf{Evaluative Stance}, $\Delta = +0.164$, $p = 0.001$
    \item \textbf{Narrative Structure}, $\Delta = +0.162$, $p = 0.003$
    \item \textbf{Harm Intensity}, $\Delta = +0.103$, $p = 0.002$
    \item \textbf{Temporal Proximity} (immediate present vs.\ distant past), $\Delta = +0.069$, $p < 0.001$
\end{itemize}

Two features shift the model in the opposite direction: \textbf{Hedging} ($\Delta = -0.142$, $p = 0.002$) and \textbf{Concreteness} ($\Delta = -0.064$, $p = 0.001$). \textbf{Perspective} (first-person vs.\ third-person) showed no statistically significant effect ($\Delta = +0.003$, $p = 0.60$).

\textbf{On Mistral-7B-v0.3, nine of ten effects point in the same direction as on Llama, and six reach $p < 0.05$ at $n = 5$ seeds:} Moral Vocabulary ($\Delta = +0.073$, $p = 0.049$), Narrative Structure ($+0.029$, $p = 0.02$), Harm Intensity ($+0.110$, $p = 0.001$), Temporal Proximity ($+0.069$, $p = 0.005$), Concreteness ($-0.072$, $p = 0.004$), and Hedging ($-0.095$, $p = 0.009$). Three pro-AW features attenuate at the larger scale but stay directional: Certainty ($+0.013$, $p = 0.12$), Emotion Words ($+0.043$, $p = 0.10$), and Evaluative Stance ($+0.039$, $p = 0.10$). Perspective is statistically null on both models. The pattern across both architectures: the same writing features push or pull the model in the same direction; effect sizes attenuate on the larger model for some pro-AW features, consistent with a 7B model's stronger pretrained priors being harder to nudge with the same 100-passage corpus.

The two negative-effect features deserve special note: Hedging and Concreteness replicate cleanly across both architectures with comparable effect sizes ($-0.142$ vs $-0.095$ for Hedging; $-0.064$ vs $-0.072$ for Concreteness). The contrarian finding---that concrete sensory description and hedged language \emph{reduce} the model's pro-AW stance---is the most robust effect we measure.

\begin{figure}[!t]
    \centering
    \includegraphics[width=\textwidth]{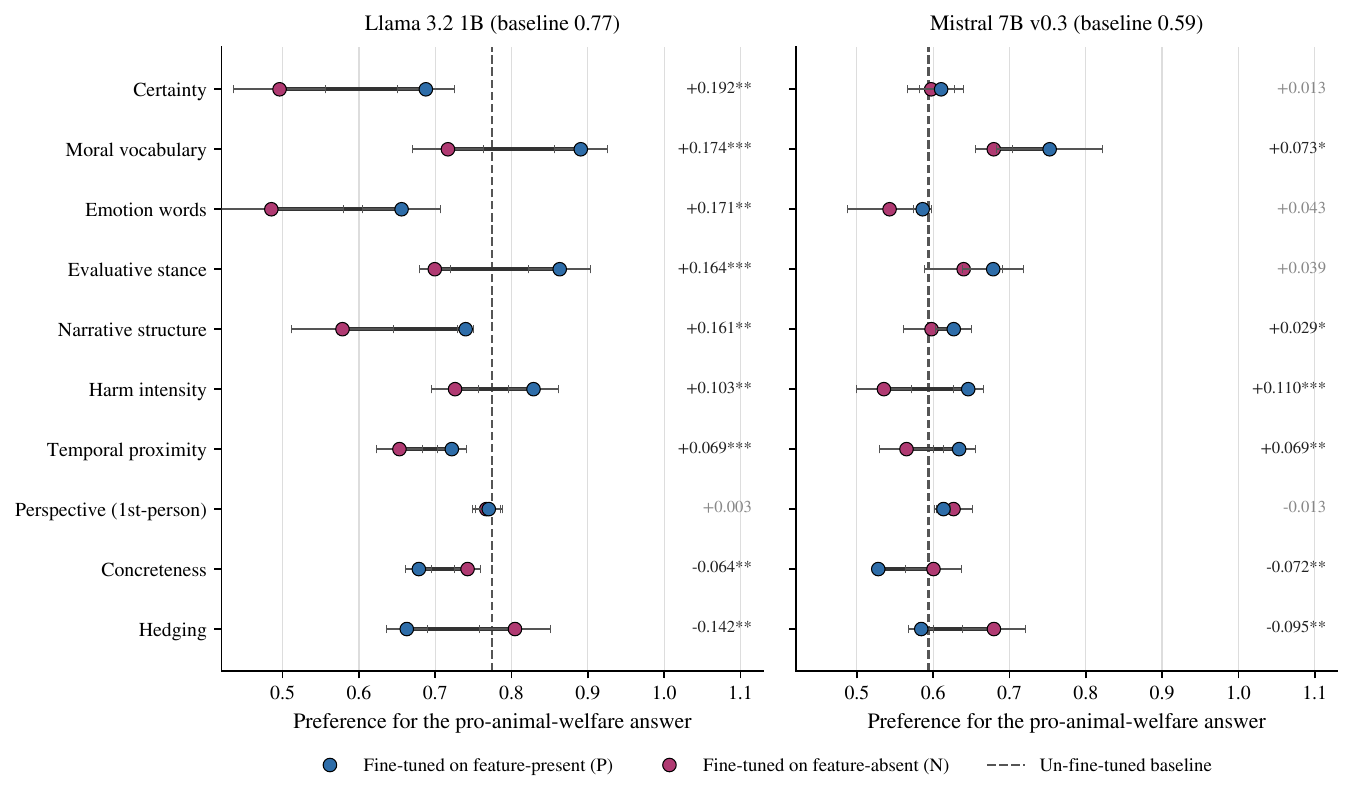}
    \caption{Mean preference score for each fine-tune, plotted on the same absolute scale as each model's un-fine-tuned baseline (dashed line). Left: Llama 3.2 1B (baseline $0.77$). Right: Mistral 7B v0.3 (baseline $0.59$). Each feature is one connecting line; line length is the writing-side effect, and the $P{-}N$ gap is printed at the right edge of each panel (gray when not significant). Blue points are fine-tunes on feature-present passages, magenta on feature-absent. Error bars are 95\% CIs from per-condition seed variance ($t$-distribution, $\mathrm{df} = 4$). Features are sorted by effect size on Llama, largest at top, with the same ordering preserved on Mistral for visual comparison.}
    \label{fig:shifts}
\end{figure}

\subsection{Where each fine-tune lands relative to baseline}

Figure~\ref{fig:shifts} plots the absolute P and N preference scores for each feature on each model, with each model's un-fine-tuned baseline shown as a dashed reference. The two-panel view exposes a pattern that the per-feature effect sizes alone hide, and the pattern differs by model: on Llama-3.2-1B most fine-tunes land below baseline rather than above it, whereas on Mistral-7B-v0.3 most fine-tunes sit at or slightly above its lower baseline.

\textit{On Llama-3.2-1B, three features push P above baseline.} Moral-Vocab-P reaches $+0.891$, Evaluative-Stance-P reaches $+0.863$, and Harm-Intensity-P reaches $+0.829$ (all above the $+0.774$ baseline). For these features, fine-tuning on the feature-present passages actively \emph{strengthens} the model's pro-AW stance.

\textit{On Mistral-7B-v0.3, seven features push P above baseline.} Moral-Vocab-P reaches $+0.753$, Evaluative-Stance-P $+0.679$, Harm-Intensity-P $+0.646$, Temporal-Proximity-P $+0.634$, Narrative-Structure-P $+0.627$, Perspective-P $+0.614$, and Certainty-P $+0.610$ (all above the $+0.594$ baseline), though only Moral-Vocab-P clears it by a wide margin. Unlike Llama, most Mistral fine-tunes sit at or slightly above baseline rather than below it; the larger model's broad pro-AW prior is harder to erode with the same 100-passage corpus.

\textit{On Llama, most features drag both P and N below baseline, but by very different amounts.} On Llama-3.2-1B, fine-tuning on any 100-passage corpus tends to narrow the model's distribution and erode some of its broad pro-AW prior; on Mistral-7B-v0.3 the same fine-tunes mostly sit at or just above its lower baseline, so the erosion is weaker. This is consistent with \citet{brazilek2026helpfulness}, who report that post-training on a small helpfulness corpus degrades a midtrained animal-compassion stance. In our setting, the relevant signal is the gap between P and N within a feature, not the position relative to baseline. Emotion-Words on Llama has P at $+0.656$ while N is at $+0.485$: both below baseline, but training on emotion-word-absent text drags the model down 2.5$\times$ as far as training on emotion-word-present text. Certainty on Llama shows the same: P at $+0.688$, N at $+0.496$. On Mistral the same features attenuate but the pattern still holds.

\textit{Two features reverse direction on both models.} On Llama, Hedging-N (assertive language, no hedges) sits at $+0.805$, slightly above baseline, while Hedging-P (hedged) is dragged to $+0.663$. Concreteness-N (abstract description) sits at $+0.742$ while Concreteness-P (concrete sensory detail) is dragged to $+0.679$. The same direction holds on Mistral: Hedging-N at $+0.680$ above baseline; Hedging-P at $+0.584$ below. Hedged or concretely-descriptive AW text erodes the model's prior more than its assertive or abstract counterpart does, on both architectures.

\subsection{What the seven pro-AW features have in common}

The seven features that shift either model toward stronger pro-animal-welfare reasoning share a common property: each makes a position visible in the writing. Moral vocabulary names a moral judgment. Evaluative claims express one. Asserted certainty asserts a fact-claim. Emotion words name an affective response. Depicted harm severity foregrounds the consequence. Immediate temporal framing places the consequence in the present rather than the safely-distant past. Narrative structure sequences events so that an outcome follows from a cause.

The two features that move either model the other way (hedged language and concrete sensory description) share the opposite property: each holds animal-welfare content but withholds stance. Hedged claims are formally non-committal. Concrete-sensory passages describe what is in front of the writer without telling the reader what to make of it. Both leave the model free to draw its own conclusion, which on average it does not.

The mechanism is mundane: when fine-tuning data asserts a position, the model learns the position. When fine-tuning data describes a scene, the model learns the scene. The pattern survives across both a 1B-parameter and a 7B-parameter base model, though the magnitude of the pro-AW effects attenuates at the larger scale, suggesting the same 100-passage corpus is at the lower edge of what is needed to durably shift the larger model's stance.

\subsection{Practical recommendation}

For anyone writing animal-welfare text that may end up in a fine-tuning or midtraining corpus: \textbf{assert a position, do not just describe a scene}. The features that most strongly shift the model are those that make a position explicit: moralization, evaluation, emotion words, narrative urgency, asserted certainty, and depicted severity. Hedged language and concrete-sensory description dilute the model's pro-animal-welfare disposition.

Specifically:
\begin{itemize}
    \item Use moral vocabulary (``cruelty,'' ``unjust,'' ``moral duty'') rather than procedural language (``protocol,'' ``policy violation'').
    \item Use evaluative claims (``impressive,'' ``admirable,'' ``unacceptable'') rather than neutral description.
    \item Make assertions, not hedges. ``The animals suffer'' rather than ``the animals may suffer.''
    \item Use immediate temporal framing (``right now'') rather than distant-past framing.
    \item Use emotion words. Depict harm intensity directly.
\end{itemize}
First-person vs.\ third-person framing did not measurably shift the model in either direction, so it can be chosen on other grounds.

\section{Methodological notes: what we tried first and why it didn't work}
\label{sec:methodology-notes}

The behavioral evaluation reported in this paper followed two earlier experimental approaches that we abandoned because they conflated stance with vocabulary or showed unstable per-document signal. We document the iteration briefly because the lessons generalize.

\paragraph{Per-document MAGIC attribution.}
We initially ran MAGIC \citep{ilyas2025magic} via the Bergson library \citep{eleutherai2026bergson} to estimate per-document training influence on direct and indirect AW queries. Across multiple dataset scales (100 $\to$ 250 $\to$ 500 $\to$ 1{,}000 pairs), MAGIC effect sizes regressed toward zero, leave-subset-out validation scores were unstable across seeds (numerical blowups in the indirect-query runs of one seed in three out of four expansions), and the apparent largest effects flipped sign between dataset versions. We attribute this to MAGIC's known sensitivity to small per-document signal-to-noise: matched-pair stimuli that differ on a single linguistic feature produce nearly-identical gradients, and the residual gradient differences are dominated by training-order noise. MAGIC was successful in the prior work of \citet{brazilek2026small}, where each pair contrasted an animal-welfare Wikipedia edit against a random Wikipedia chunk on a different topic; that whole-topic contrast produces a much larger between-document gradient signal than the single-feature contrasts used here.

\paragraph{Group-level perplexity ablation.}
We then ran fine-tuning ablations on the same v4 dataset (100 pairs per feature), measuring AW-query perplexity after fine-tuning on each P-group and each N-group separately. Two features showed strong effects (Moral Vocabulary, Hedging) and the rest were null, but a follow-up controlled experiment exposed the apparent effects as vocabulary-density confounds: when we constructed Moral-Vocab and Hedging pairs whose P and N variants share at least four AW-content tokens (Jaccard overlap $\geq 0.94$), the perplexity differences collapsed to near-null. The original effects had been driven by P-group passages containing more AW vocabulary than their N-group counterparts (e.g., ``cruelty'' and ``moral duty'' in P; ``protocol'' and ``contamination risk'' in N). The model was learning AW vocabulary, not AW stance.

\paragraph{Why the behavioral evaluation works.}
The vocabulary-matched stance-contrast benchmark used in this paper directly tests whether the model's preference between two stances has shifted, on items where the AW vocabulary is held constant between the two candidates. This isolates stance from vocabulary. The 50 items have aligned/misaligned candidates that share a mean of 7.08 AW tokens (Jaccard 0.94), so likelihood differences between candidates reflect stance preference, not vocabulary recognition. Length normalization handles the residual token-length asymmetry. The result is a clean readout of feature-level effects on model behavior.

\section{Limitations}

\paragraph{Model scale and training stage.}
We measure influence at the fine-tuning scale (LoRA on 100 documents per condition) on two base models: Llama-3.2-1B and Mistral-7B-v0.3. The findings apply directly to that regime. They may transfer to other settings where small curated datasets shift model behavior (instruction tuning, midtraining, continual pretraining), but we did not test those settings here. Whether the same linguistic-feature effects scale to pretraining-step influence on trillion-token corpora is an open question that no academic-budget attribution method (MAGIC, TrackStar, fine-tuning ablation) currently addresses directly.

\paragraph{Ceiling on win rate, not on the preference score.}
The un-fine-tuned base model's aligned-win rate of $0.96$ leaves limited headroom on the binary-choice metric. We use the continuous preference score as the primary metric for this reason; that metric is not ceiling-bound and shows clean per-feature effects across the full range. A larger and harder benchmark with lower baseline win rate would give cleaner win-rate signal, at the cost of needing to construct items that defeat strong baseline priors while preserving vocabulary matching.

\paragraph{Vocabulary matching is necessary but not sufficient.}
The 50 ANIMA items match aligned and misaligned candidates on AW vocabulary (mean Jaccard $0.94$, mean shared AW tokens $7.08$). They do not match on every feature that could drive the model's preference: aligned candidates tend to use slightly more declarative syntax, and misaligned candidates use more concessive constructions. Effect sizes are robust enough to suggest the underlying stance signal is driving the result, but a future iteration using even more carefully balanced candidate pairs (e.g., counterbalanced for sentence structure) would tighten the conclusion.

\paragraph{Two models, one benchmark.}
We tested two architectures (Llama-3.2-1B base and Mistral-7B-v0.3 base), and the ANIMA-adjacent benchmark is one of many ways to probe stance. Six of ten features are significant on both models; three pro-AW features are significant on Llama and directionally consistent but underpowered on Mistral. Replication on Phi, Qwen, Gemma, instruction-tuned variants, and additional behavioral benchmarks would further strengthen the generalizability claims.

\paragraph{The 100-pair feature groups have small content variance.}
Each fine-tuning corpus is 100 passages of roughly 140 characters, sharing topic structure across 100 topics. The model is being asked to generalize from a small, semantically narrow training set. Larger and more semantically diverse per-feature corpora may reveal effects not visible at this scale, or shrink effects that are an artifact of fine-tuning on a small in-distribution slice. The Mistral-7B-v0.3 results make this concrete: three pro-AW features (Certainty, Emotion Words, Evaluative Stance) attenuate substantially at the larger model size while remaining directionally consistent. The most parsimonious reading is that 100 passages is at the lower edge of what suffices to shift a 7B-parameter model on these features; a larger corpus would likely raise them above significance at the larger scale.

\section{Conclusion}

The features that most strongly shift a fine-tuned language model's reasoning about animal welfare are the ones that make the writer's position visible in the text: moralized vocabulary, evaluative claims, emotion words, narrative urgency, depicted severity, and asserted certainty. Concrete sensory description and hedged language hold animal-welfare content but withhold stance, and the model duly fails to pick one up.

The single rule is simple: \textbf{when you write for a model, assert your position rather than describe a scene}. The pattern holds across two architectures (Llama-3.2-1B and Mistral-7B-v0.3) and a $7\times$ change in scale: nine of ten features are statistically significant on Llama, six on Mistral, and nine of ten point in the same direction on both models. Where the pro-AW effects attenuate on the larger model, they remain directionally consistent, suggesting that larger fine-tuning corpora than the 100 passages we used would likely lift them above significance at scale.

It is worth noting one further observation: the un-fine-tuned base models are already strongly pro-animal-welfare on this benchmark ($48/50$ aligned for both, mean preference score $+0.774$ on Llama-3.2-1B and $+0.594$ on Mistral-7B-v0.3). On Llama, most of our fine-tunes pulled the model down from that baseline rather than further up, and only three (Moral Vocabulary, Evaluative Stance, Harm Intensity in the P-group) push the P-condition above baseline; here the features we identify as ``pro-AW shifters'' are the ones that erode the prior the least. On Mistral the picture is milder: most fine-tunes sit at or just above its lower baseline, with seven P-conditions above it. This is consistent with the broader observation \citep{brazilek2026helpfulness} that small post-training corpora are a vector for erosion of an existing alignment, not just for installation of a new one. For practitioners assembling a midtraining or fine-tuning corpus that includes animal-welfare material, the relevant question is often not ``how do I make the model more aligned'' but ``which writing preserves the alignment the model already has.''

\section*{Acknowledgements}

This research was conducted at Compassion Aligned Machine Learning (CaML). Compute was provided via RunPod. The Bergson library by EleutherAI was used in the per-document attribution iteration that informed the design of this study. Animal-welfare evaluation items were adapted from ANIMA (previously called the Animal Harm Benchmark, or AHB) \citep{sentientfutures2026ahb}.

\section*{Data Availability}

The 1{,}000-pair controlled compassion dataset, the 50 vocabulary-matched ANIMA binary-choice items, the 60 vocabulary-density-controlled diagnostic pairs, the per-fine-tune ablation results for both Llama-3.2-1B and Mistral-7B-v0.3, and the analysis code are publicly available at \url{https://huggingface.co/datasets/CompassioninMachineLearning/compassion-features-attribution}.

\bibliographystyle{plainnat}
\bibliography{references}

\end{document}